\documentclass[conference]{IEEEtran}
\IEEEoverridecommandlockouts
\usepackage{amsmath,amssymb,amsfonts}
\usepackage{algorithmic}
\usepackage{graphicx}
\usepackage{textcomp}
\usepackage{xcolor}
\def\BibTeX{{\rm B\kern-.05em{\sc i\kern-.025em b}\kern-.08em
    T\kern-.1667em\lower.7ex\hbox{E}\kern-.125emX}}

\usepackage[
backend=biber,
style=numeric,
sorting=ynt
]{biblatex}
\usepackage{caption}
\usepackage{subcaption}
\usepackage{multirow}

\usepackage{fancyhdr}

\fancypagestyle{copyright}{%
    \fancyhf{}
    \cfoot{\footnotesize%
        \textcopyright~978-1-6654-8045-1/22/\$31.00 ©2022 IEEE
    }
}

\begin{document}

\title{Deep Semantic Model Fusion for Ancient Agricultural Terrace Detection
}

\author{\IEEEauthorblockN{Yi Wang$^{1,2}$, Chenying Liu$^{1,2}$, Arti Tiwari$^{3}$, Micha Silver$^{3}$, Arnon Karnieli$^{3}$, Xiao Xiang Zhu$^{1}$, Conrad M Albrecht$^{2}$}
\IEEEauthorblockA{\textit{$^1$Chair of Data Science in Earth Observation, Technical University of Munich (TUM), Germany} \\
\textit{$^2$Remote Sensing Technology Institute, German Aerospace Center (DLR), Germany} \\
\textit{$^3$The Remote Sensing Laboratory, Institutes for Desert Research, Ben Gurion University (BGU), Israel}}
}

\maketitle

\begin{abstract}
Discovering ancient agricultural terraces in desert regions is important for the monitoring of long-term climate changes on the Earth's surface. However, traditional ground surveys are both costly and limited in scale. With the increasing accessibility of aerial and satellite data, machine learning techniques bear large potential for the automatic detection and recognition of archaeological landscapes. In this paper, we propose a deep semantic model fusion method for ancient agricultural terrace detection. The input data includes aerial images and LiDAR generated terrain features in the Negev desert. Two deep semantic segmentation models, namely DeepLabv3+ and UNet, with EfficientNet backbone, are trained and fused to provide segmentation maps of ancient terraces and walls. The proposed method won the first prize in the International AI Archaeology Challenge. Codes are available at https://github.com/wangyi111/international-archaeology-ai-challenge.
\end{abstract}

\thispagestyle{copyright}


\section{Introduction}

Discovering ancient agricultural terraces in desert regions has great importance for both archaeological and anthropological research in the monitoring of long-term climate change. First, the information can be gathered to advance our knowledge of ancient human endeavors. Second, indicating the potential use of surface runoff may also help reveal locations for enhancing future food production. While there is a growing need to discover new agricultural land resources, traditional ground surveys are limited in scale. With the development of modern imaging and machine learning techniques, to discover these regions on a large scale becomes possible \cite{bickler2021machine}.

\begin{figure}
    \centering
    \includegraphics[width=\linewidth]{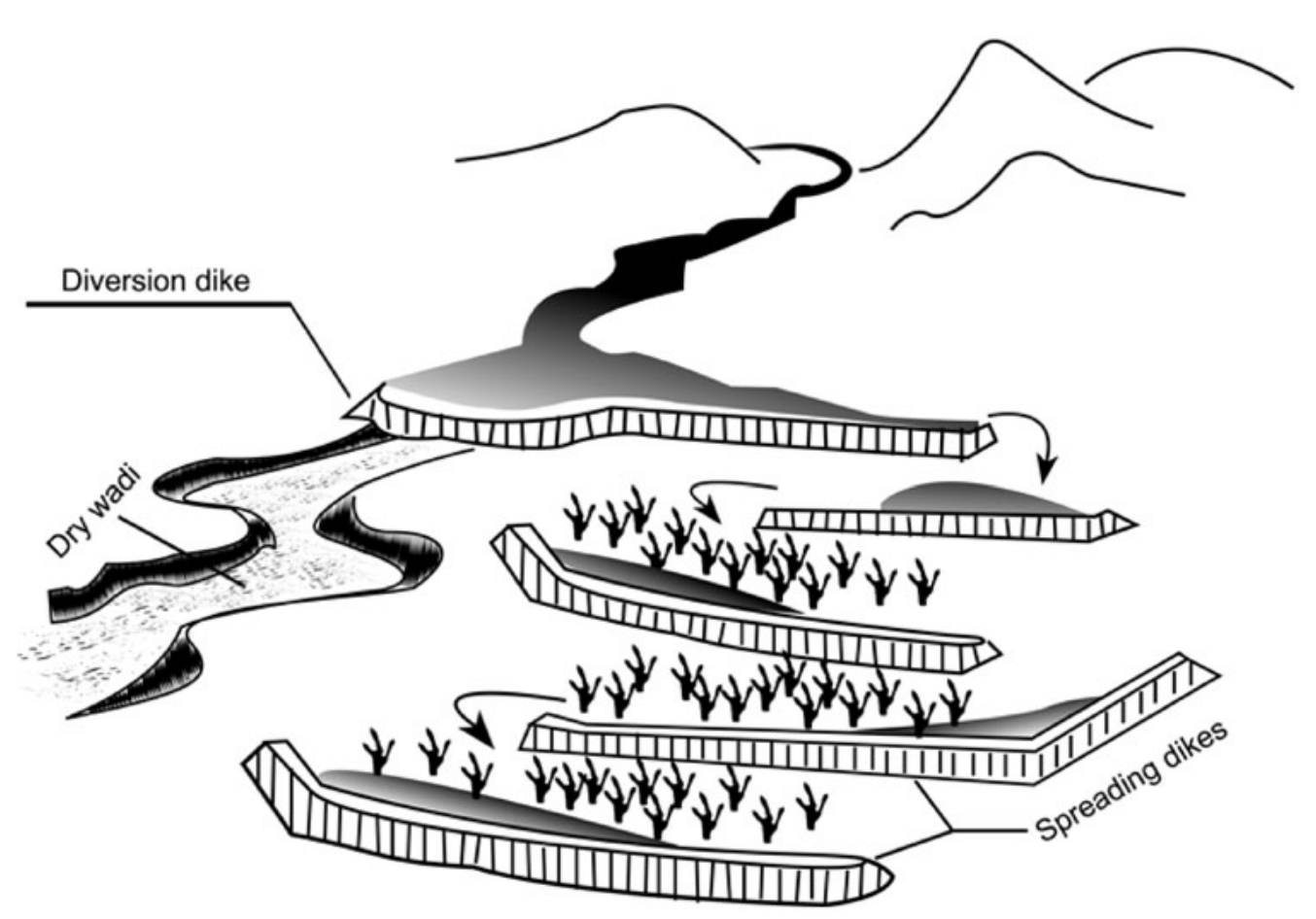}
    \caption{A typical flood water harvesting system (figure adapted from \cite{french1964water,rango2009water}).}
    \label{fig:harvesting}
\end{figure}

The Negev, Israel, is a subtropical desert with occasional precipitation in the autumn, winter, and spring, while the hot summer is completely dry for almost half a year from about May to October. Thousands of ancient dry stonewalls (named terraces) were built in the central Negev Highlands during ancient times, mainly between the 4th and the 7th centuries, across ephemeral stream channels (wadis, Figure \ref{fig:harvesting}). This water harvesting technique is very common in wadi beds with gentle slopes. As a result of the slow water velocity, eroded sediments and nutrients usually settle in the wadi bed and create good agricultural land. 

\begin{figure*}
     \centering
     \begin{subfigure}[b]{0.3\textwidth}
         \centering
         \includegraphics[width=\textwidth]{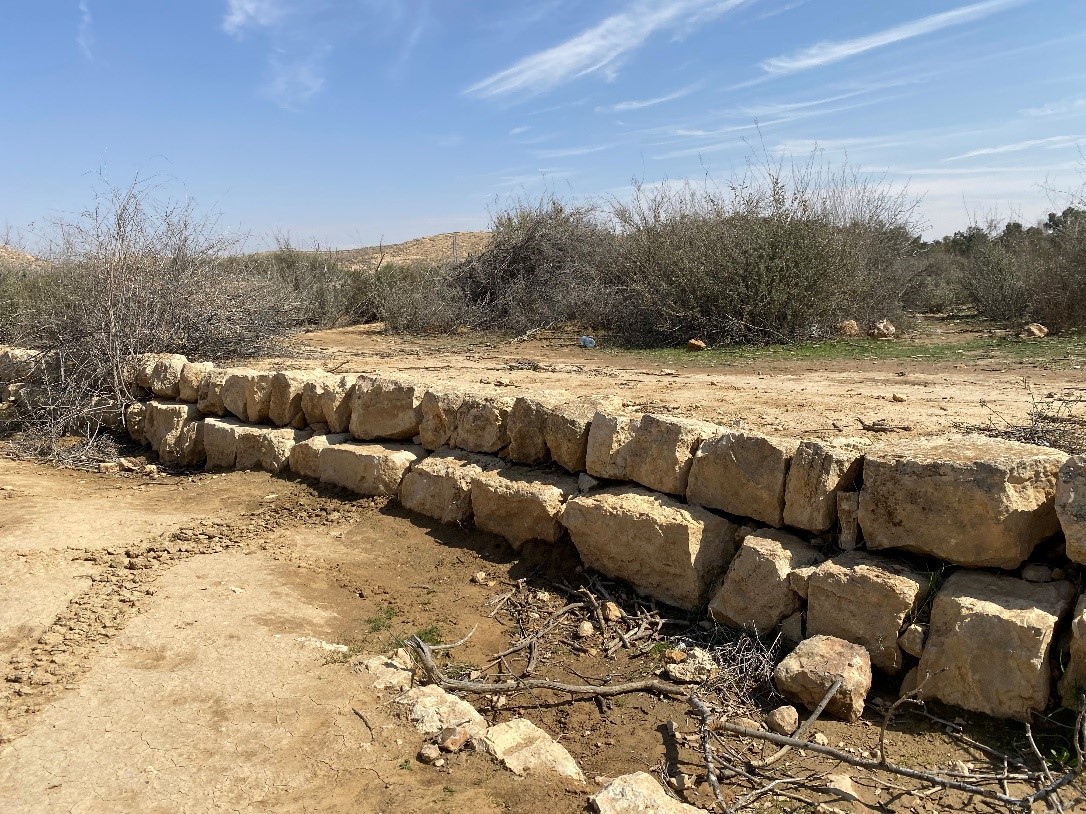}
         \caption{Example exposed terrace.}
     \end{subfigure}
     \begin{subfigure}[b]{0.3\textwidth}
         \centering
         \includegraphics[width=\textwidth]{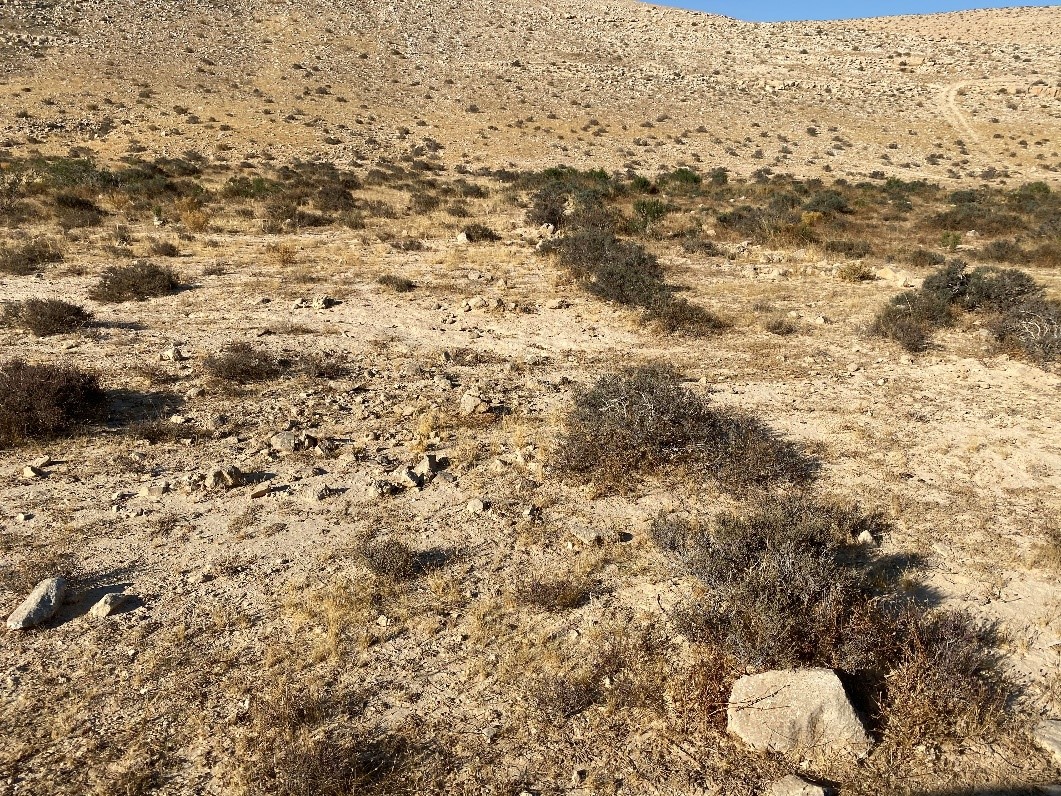}
         \caption{Example buried terrace.}
     \end{subfigure}
     \begin{subfigure}[b]{0.3\textwidth}
         \centering
         \includegraphics[width=\textwidth]{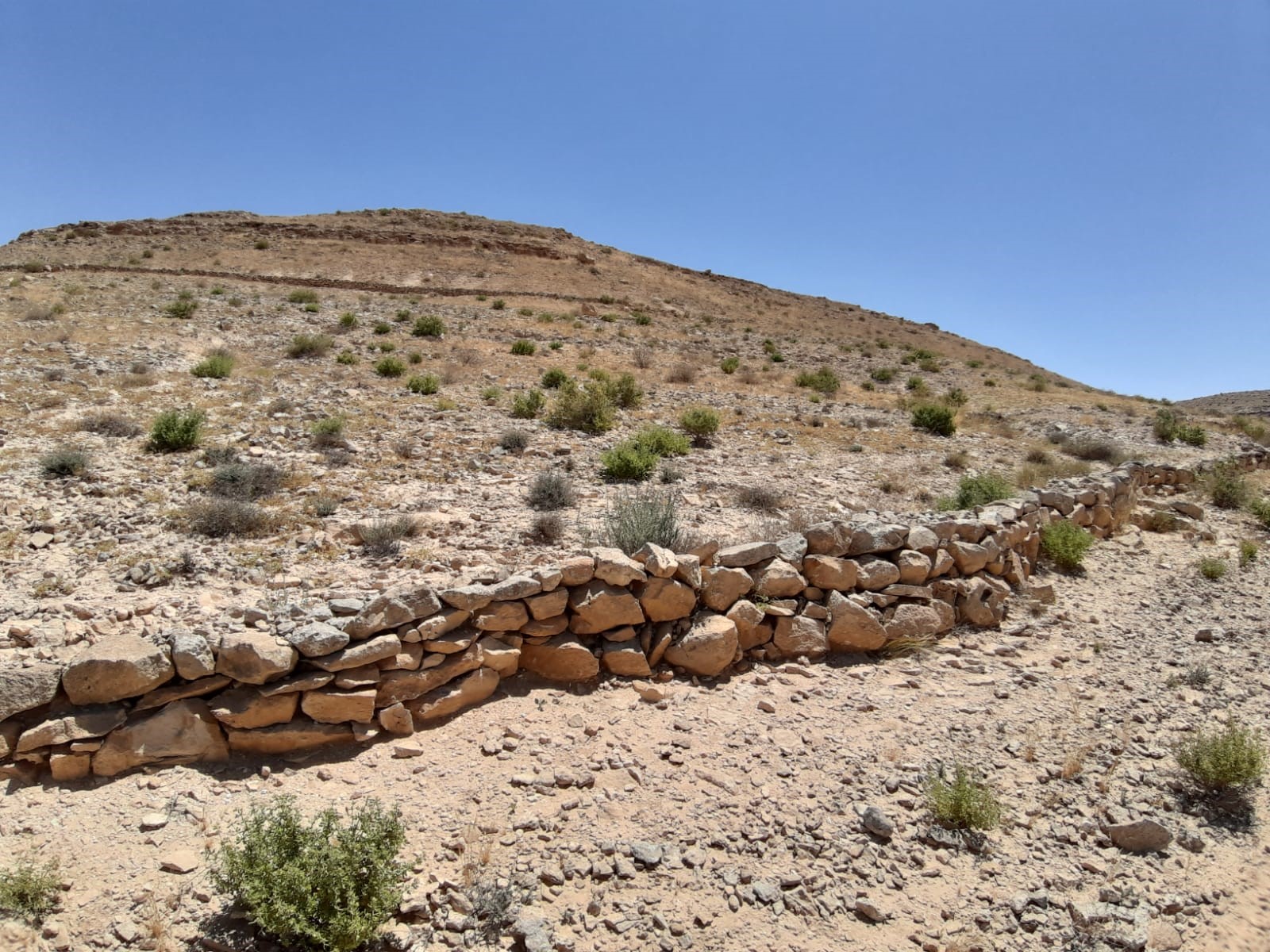}
         \caption{Example stonewall fence.}
     \end{subfigure}
        \caption{Ground photos of example terraces and walls \cite{internationalaiarcheology}.}
        \label{fig:ground-photo}
\end{figure*}

The ancient agricultural terraces in the Negev desert have been abandoned since the 7th century, but many stone terrace walls are still intact.  Nowadays, the terraces can be observed from the surface either as exposed stonewalls or as buried stonewalls that can be identified by the above-ground vegetation (Figure \ref{fig:ground-photo}).
In the International AI Archeology Challenge \cite{internationalaiarcheology}, aerial image data including RGB images and terrain feature images generated from LiDAR surveys are provided to detect ancient terraces and walls as a multiclass semantic segmentation task.

In this paper, we describe the winning solution of the International AI Archeology Challenge. Specifically, we propose a deep semantic model fusion method that combines the soft prediction score of DeepLabv3+ \cite{chen2018encoder} and U-Net \cite{ronneberger2015u} to segment the ancient agricultural terraces and walls. Our results verify the promising potential of deep neural networks in the application of archeological landscape recognition.




\section{Methodology}
In this section, we describe the details of the proposed deep semantic model fusion method. We first separately train a U-Net and a DeepLabv3+ model, each with EfficientNet \cite{tan2019efficientnet} as the encoder backbone. During inference, the soft prediction scores from each model are fused together to decide the final output class.

\subsection{Semantic segmentation models}

\subsubsection{U-Net}

\begin{figure}
    \centering
    \includegraphics[width=\linewidth]{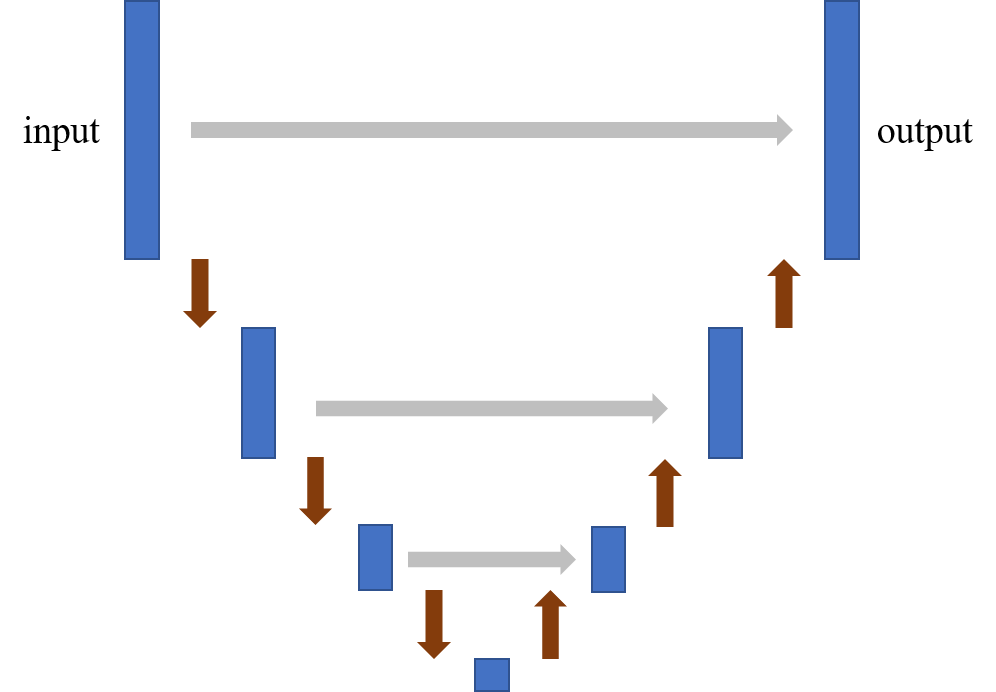}
    \caption{U-Net \cite{ronneberger2015u}. Multi-scale features from the downsampling encoder are concatenated to those from the upsampling decoder. Each blue box corresponds to a multi-channel feature map.
}
    \label{fig:unet}
\end{figure}

The U-Net \cite{ronneberger2015u} architecture was initially designed for biomedical image segmentation. As is shown in Figure \ref{fig:unet}, the network consists of a contracting path and an expansive path, which gives it the U-shaped architecture. The downsampling encoder path is used to capture the multiscale context in the image, which is originally a stack of convolutional and max pooling layers. The upsampling decoder path is the symmetric expanding path, which is used to enable precise localization using transposed convolutions. Multiscale features from each layer of the encoder are kept and concatenated to the corresponding decoder layer, allowing the network to propagate context information to higher resolution layers.

\subsubsection{DeepLabv3+}

\begin{figure}
    \centering
    \includegraphics[width=0.8\linewidth]{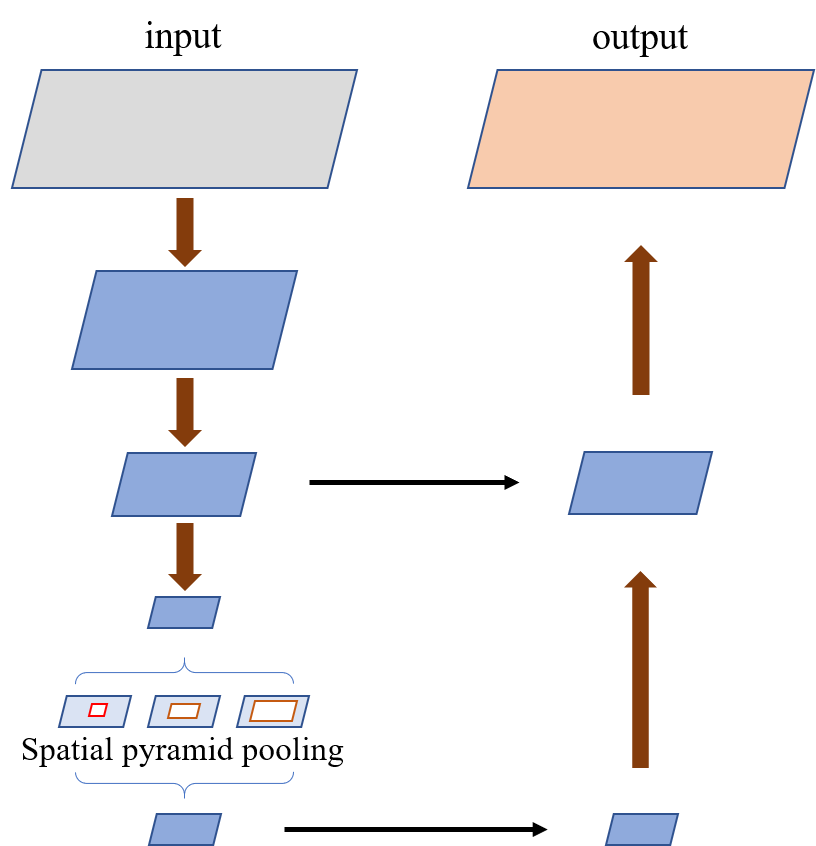}
    \caption{DeepLabv3+ \cite{chen2018encoder}. The encoder contains a spatial pyramid pooling module \cite{chen2017rethinking} that allows extracting features at an arbitrary resolution by applying atrous convolution. The simple decoder helps to recover detailed object boundaries.}
    \label{fig:deeplabv3plus}
\end{figure}

The DeepLabv3+ \cite{chen2018encoder} architecture is based on the DeepLab semantic segmentation model series \cite{chen2014semantic,chen2017deeplab,chen2017rethinking}. As is shown in Figure \ref{fig:deeplabv3plus}, the network has an encoder-decoder structure. Atrous Spatial Pyramid Pooling \cite{chen2017deeplab} is used in the encoder to capture multi-scale contextual information. Parallel atrous convolution with different rates is applied in the input feature map, and fused together. Depthwise separable convolution, or atrous separable convolution, helps to
reduce the computation complexity while maintaining similar (or better) performance. A simple decoder is used to recover detailed spatial information.

\begin{figure*}
     \centering
     \begin{subfigure}[b]{0.18\textwidth}
         \centering
         \includegraphics[width=\textwidth]{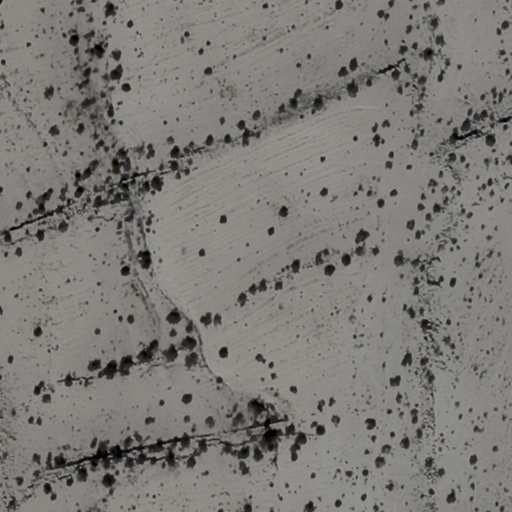}
         \caption{Ortho.}
     \end{subfigure}
     \begin{subfigure}[b]{0.18\textwidth}
         \centering
         \includegraphics[width=\textwidth]{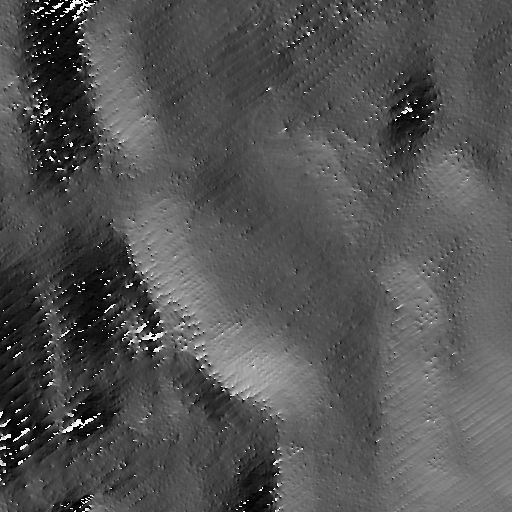}
         \caption{Aspect.}
     \end{subfigure}
     \begin{subfigure}[b]{0.18\textwidth}
         \centering
         \includegraphics[width=\textwidth]{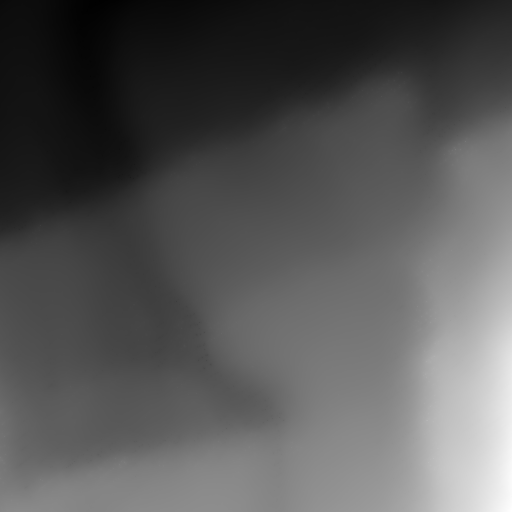}
         \caption{DTM.}
     \end{subfigure}
     \begin{subfigure}[b]{0.18\textwidth}
         \centering
         \includegraphics[width=\textwidth]{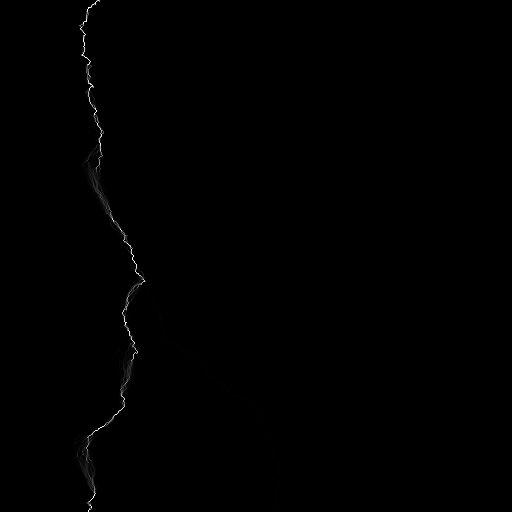}
         \caption{Flowacc.}
     \end{subfigure}
     \begin{subfigure}[b]{0.18\textwidth}
         \centering
         \includegraphics[width=\textwidth]{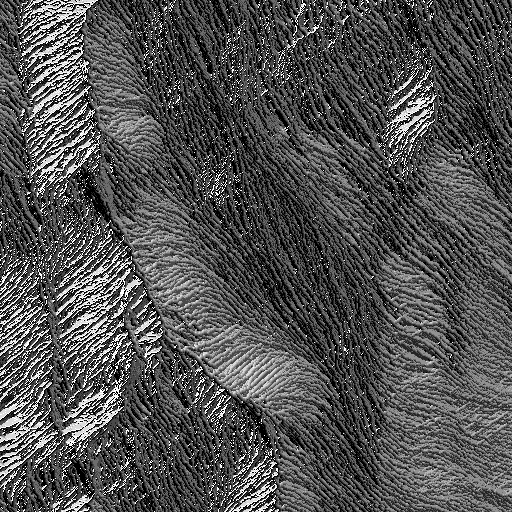}
         \caption{Flowdir.}
     \end{subfigure}
     \begin{subfigure}[b]{0.18\textwidth}
         \centering
         \includegraphics[width=\textwidth]{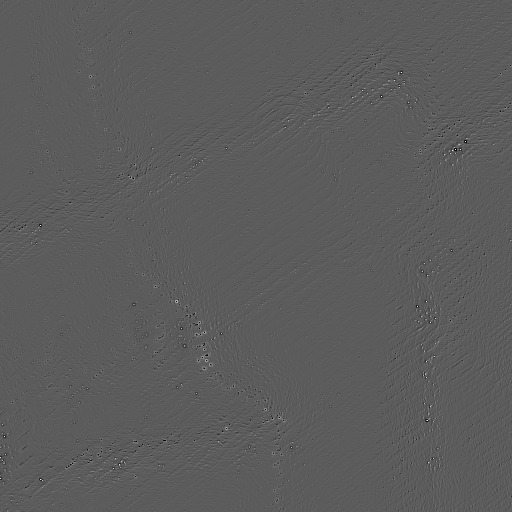}
         \caption{Pcurv.}
     \end{subfigure}
     \begin{subfigure}[b]{0.18\textwidth}
         \centering
         \includegraphics[width=\textwidth]{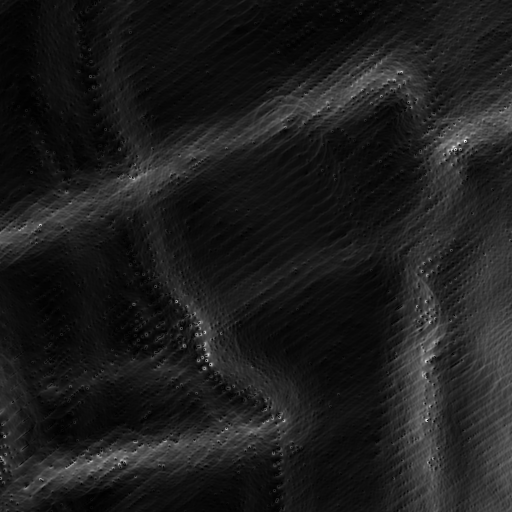}
         \caption{Slope.}
     \end{subfigure}
     \begin{subfigure}[b]{0.18\textwidth}
         \centering
         \includegraphics[width=\textwidth]{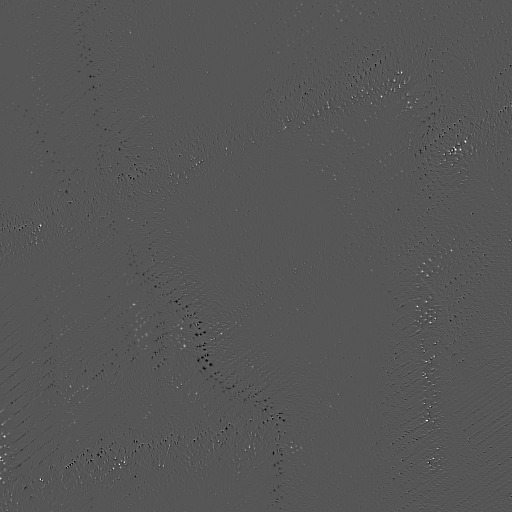}
         \caption{Tcurv.}
     \end{subfigure}
     \begin{subfigure}[b]{0.18\textwidth}
         \centering
         \includegraphics[width=\textwidth]{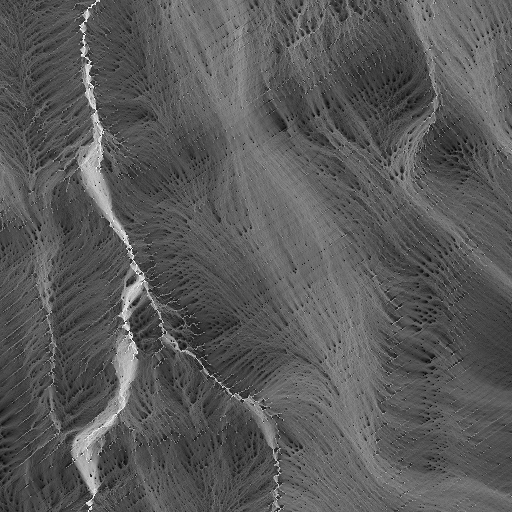}
         \caption{TWI.}
     \end{subfigure}
     \begin{subfigure}[b]{0.18\textwidth}
         \centering
         \includegraphics[width=\textwidth]{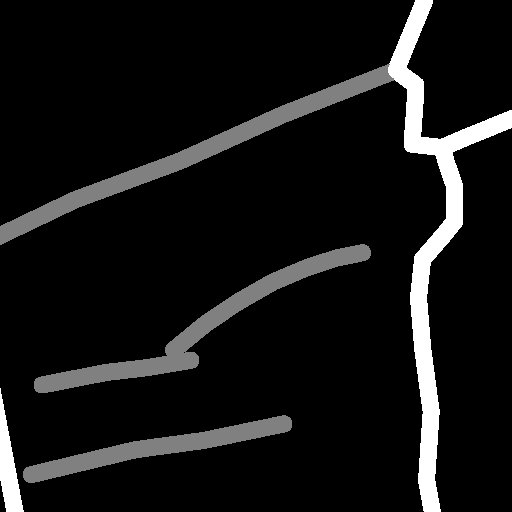}
         \caption{Ground truth.}
     \end{subfigure}

        \caption{Example training data. (a) is aerial RGB image; (b)-(i) are terrain feature images generated from LiDAR survey. (j) is the ground truth mask, where gray represents terraces and white represents walls.}
        \label{fig:data}    
\end{figure*}

\subsection{Encoder backbones}

We choose EfficientNet \cite{tan2019efficientnet} as the encoder backbones for both U-Net and DeepLabv3+. EfficientNet is a convolutional neural network architecture and scaling method that uniformly scales all dimensions of depth/width/resolution using a compound coefficient. Unlike conventional practice that arbitrarily scales these factors, the EfficientNet scaling method uniformly scales network width, depth, and resolution with a set of fixed scaling coefficients.

\subsection{Semantic Model Fusion}

After the separate training of both U-Net and DeepLabv3+, we merge the output probability maps from both models to get the final segmentation map:

\begin{equation}
\text { Out }_{(i, j)}=\alpha \cdot UNet_{(i, j)}+(1-\alpha) \cdot \text {DeepLabv3plus}_{(i, j)}
\end{equation}

\noindent where $i,j$ indicates each pixel, and $\alpha$ is a weighting parameter.

\subsection{Losses}

The model outputs 3 class probabilities: terrace, wall and background. Due to the high unbalancing of foreground (terraces and walls) and backgroud pixels, a naive cross entropy loss may push the model towards predicting only the background. To tackle this issue, we add a DICE loss that optimizes the overlap of predictions and ground truth masks for each class:

\begin{equation}
\text{DiceLoss}=1-\frac{2 T P}{2 T P+F N+F P}
\end{equation}

\noindent where TP, FN, FP represent true positive, false negative and false positive, respectively. The weighted mean of the cross entropy loss and the Dice loss is used as the total loss:

\begin{equation}
\text {TotalLoss}=\beta \cdot \text { CrossEntropyLoss }+(1-\beta) \cdot \text { DiceLoss }
\end{equation}

\subsection{Data augmentations}

The data sizes in archeology are relatively small, which may lead to strong overfitting with deep neural networks. To tackle this issue, we introduce various data augmentations to increase the diversity of the training data. Apart from commonly used RandomResizedCrop and RandomHorizontalFlip in natural images, we explicitly add more geometric and color augmentations including RandomVerticalFlip, RandomRotation, RandomAffine and RandomGaussianBlur.

\section{Experiments}

\begin{figure*}
     \centering
     \begin{subfigure}[]{0.18\textwidth}
         \centering
         \includegraphics[width=\textwidth]{figures/input/ortho_258.png}
         
         \vspace{1mm}         
         \includegraphics[width=\textwidth]{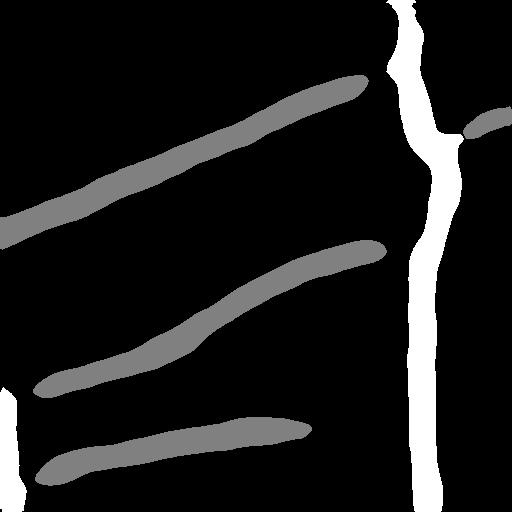}
         
         \vspace{1mm}         
         \includegraphics[width=\textwidth]{figures/input/mask_258.png}
     \end{subfigure}     
     \begin{subfigure}[]{0.18\textwidth}
         \centering
         \includegraphics[width=\textwidth]{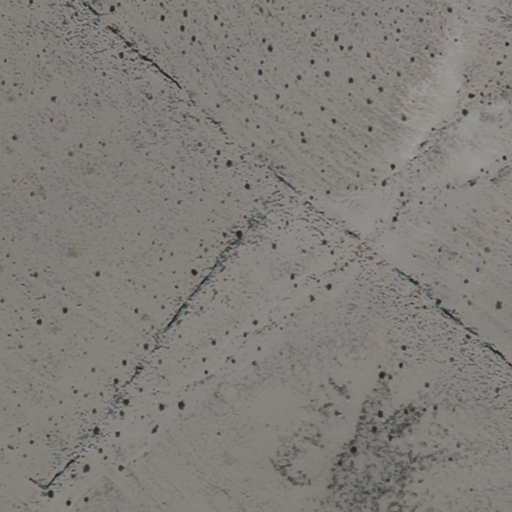}
         
         \vspace{1mm}         
         \includegraphics[width=\textwidth]{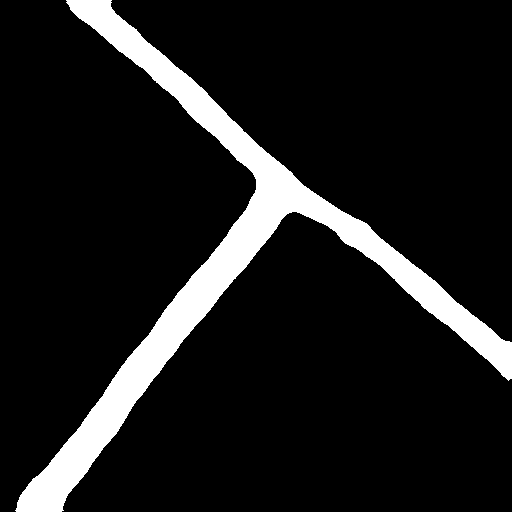}
         
         \vspace{1mm}         
         \includegraphics[width=\textwidth]{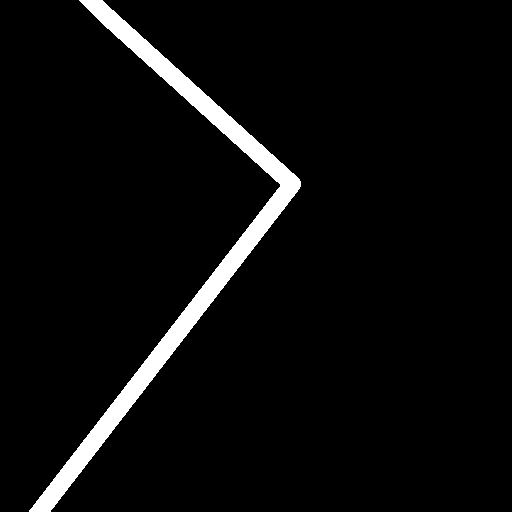}         
     \end{subfigure}
     \begin{subfigure}[]{0.18\textwidth}
         \centering
         \includegraphics[width=\textwidth]{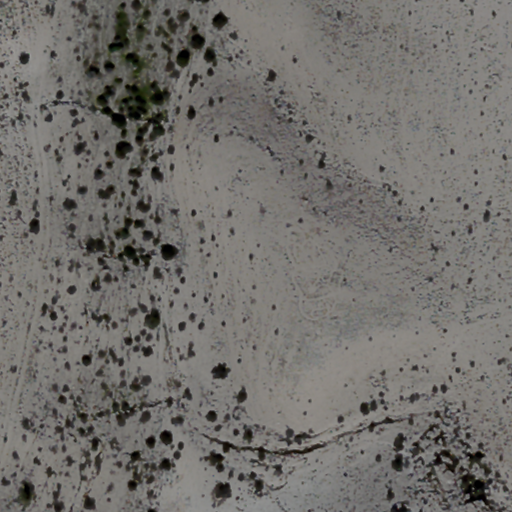}
         
         \vspace{1mm}        
         \includegraphics[width=\textwidth]{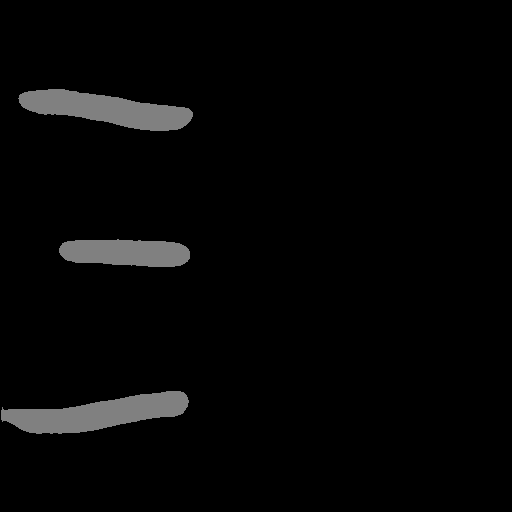}
         
         \vspace{1mm}         
         \includegraphics[width=\textwidth]{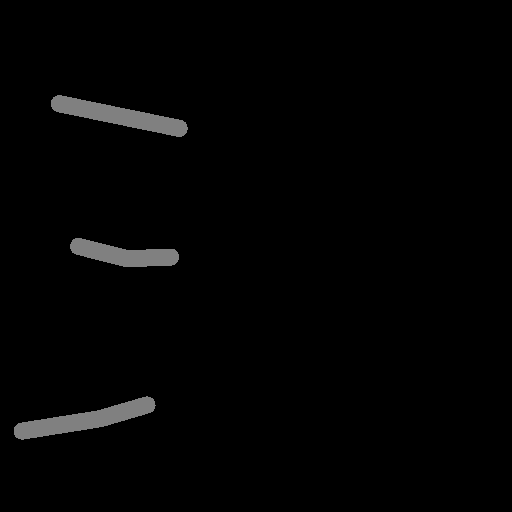}         
     \end{subfigure}
     \begin{subfigure}[]{0.18\textwidth}
         \centering
         \includegraphics[width=\textwidth]{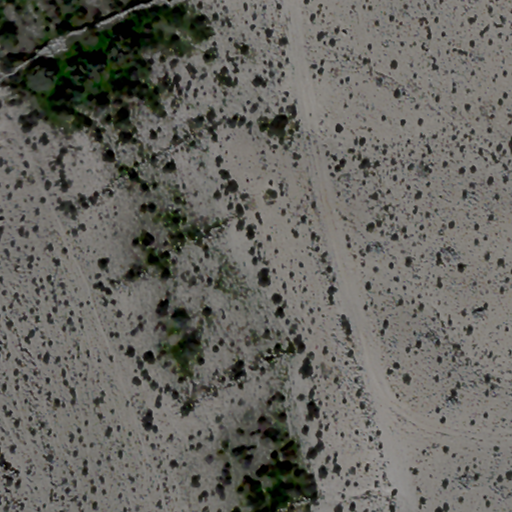}
         
         \vspace{1mm}         
         \includegraphics[width=\textwidth]{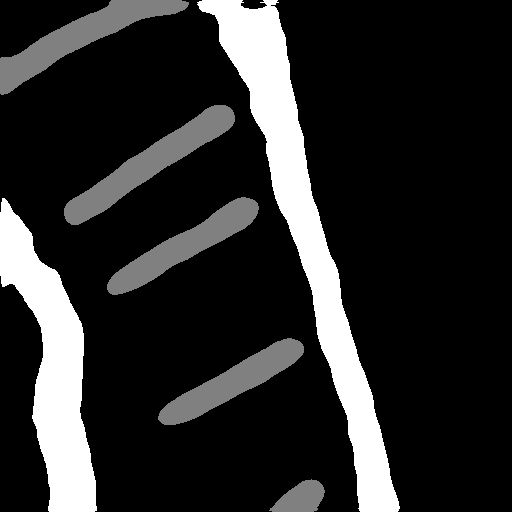}
         
         \vspace{1mm}         
         \includegraphics[width=\textwidth]{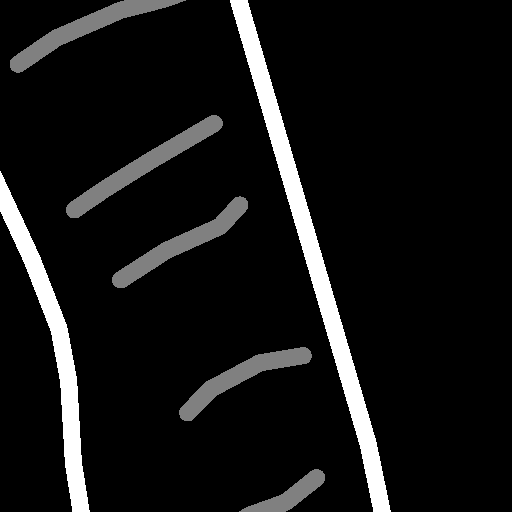}         
     \end{subfigure}
     \begin{subfigure}[]{0.18\textwidth}
         \centering
         \includegraphics[width=\textwidth]{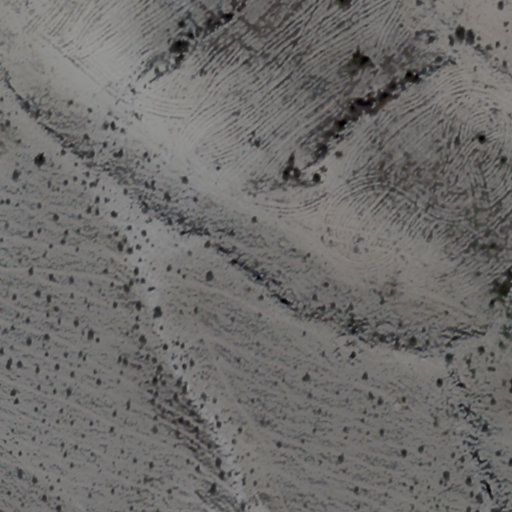}
         
         \vspace{1mm}         
         \includegraphics[width=\textwidth]{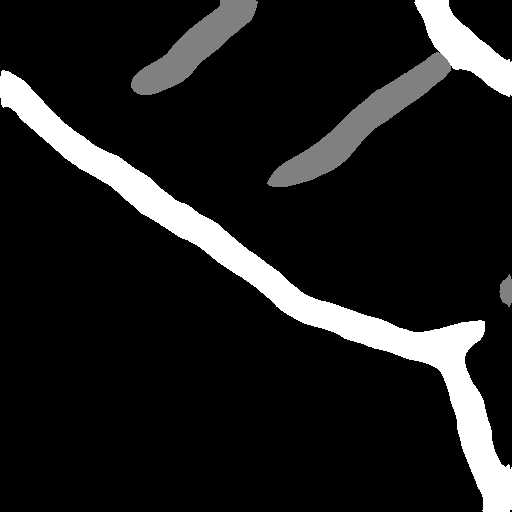}
         
         \vspace{1mm}         
         \includegraphics[width=\textwidth]{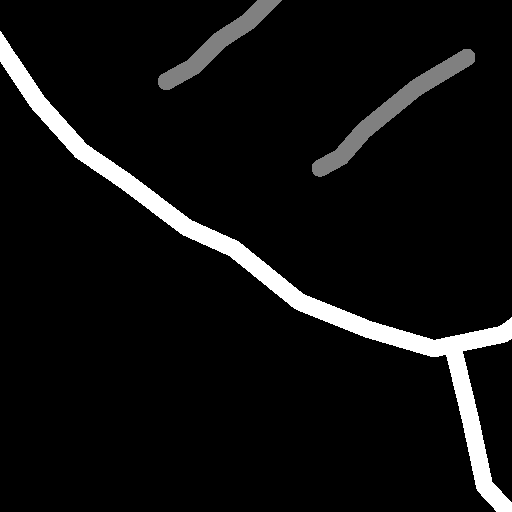}         
     \end{subfigure}
        \caption{Qualitative evaluation results of 5 example patches. The columns correspond to the 5 patches; the rows correspond to orthophotos, predicted segmentation maps and ground truth masks, respectively.}
        \label{fig:predictions}    
\end{figure*}

\begin{table*}[]
\caption{Evaluation metrics on the validation set.}
\label{tab:metrics}
\centering
\begin{tabular}{ccccccccc}
\hline
           & \multicolumn{3}{c}{terrace}                & \multicolumn{3}{c}{wall}                   & \multirow{2}{*}{IoU} & \multirow{2}{*}{mIoU} \\
           & precision    & recall       & F1           & precision    & recall       & F1           &                      &                       \\ \hline
U-Net      & \textbf{.87} & .12          & .21          & \textbf{.60} & .86          & \textbf{.71} & .45                  & .33                   \\
DeepLabv3+ & .31          & \textbf{.41} & .35          & .53          & \textbf{.90} & .67          & .44                  & .36                   \\
Fusion     & .65          & .29          & \textbf{.40} & .57          & .89          & .70          & \textbf{.47}         & \textbf{.39}          \\ \hline
\end{tabular}
\end{table*}

\subsection{Data}

The input data consists of decimeter resolution aerial RGB orthophotos and 8 terrain feature images generated from LiDAR survey. Specifically, the terrain features are Aspect, Digital Terrain Model (DTM), Flow accumulation, Flow direction, P-curve, T-curve, Slope and Topographic wetness index. The ground truth masks contain 3 classes: terrace, wall and background. Figure \ref{fig:data} shows the various features of an example patch. 500 patches with ground truth masks are provided in the training phase, and 200 patches (masks hidden) are used for testing. All images are with the size 512x512.

\subsection{Implementation Details}
We split the training data into 400/100 training/validation splits. All features (3 channel RGB + 8 channel terrain features) are used as input to the networks. 

We use ImageNet pretrained EfficientNet-B5 as the encoder backbone for both U-Net and DeepLabv3+ models. The model fusion weights $\alpha$ and the loss weighting parameter $\beta$ are both set to 0.5.

We use the batch size 16, learning rate 0.001, and AdamW optimizer. Training one model on an NVIDIA RTX 3090 takes about 1 hour.

\subsection{Results}

\subsubsection{Qualitative evaluation}

Figure \ref{fig:predictions} shows the predicted segmentation maps for 5 example patches. It can be seen that the segmentation results are very close to ground truth masks from a first look. While it is not easy for none-experts to visually check, most of the terraces and walls are detected successfully by the model. A closer look shows that the exact shape of the objects and the detailed location of the pixels are still hard to recognize. However, the situation here is different from natural images, where clear boundary information can be defined from the input data. From the aspect of ancient agricultural terraces (especially the buried ones), only the rough structures are possible to be detected even by experts.

\subsubsection{Quantitative evaluation}

The final evaluation score for the challenge is two-calss (foreground only) intersection over union (IoU):

\begin{equation}
I o U=\left|\frac{\{(i, j): P(i, j)=T(i, j)>0\}}{\{(i, j): P(i, j)>0 \text { or } T(i, j)>0\}}\right|
\end{equation}

\noindent where P,T denote the predicted map and ground truth mask. Our proposed method reached a final score of 0.31 on the hidden testing data. Besides that, we consider also precision, recall, F1 score and mean IoU (mIoU) to evaluate the prediction results. Table \ref{tab:metrics} reports the evaluation metrics on the validation set, which verify the advantage of the fusion strategy. While U-Net has a higher precision, DeepLabv3+ has a higher recall. The fused model in the end has better scores on the balanced metrics like F1 score and IoU. The general low metric values compared to the impressive qualitative results confirm the difficulty of recognizing detailed terrace boundaries from the input data. Furthermore, it can be seen that terraces are performing worse than walls, which reflects the fact that stonewalls are generally more complete and regular while terraces are usually split into pieces.

\section{Discussion}

\begin{table*}[]
\caption{Feature importance evaluation. We report the IoU scores after removing one feature during the inference, i.e., replacing the corresponding channel with zero values. The columns represent which feature to remove.}
\label{tab:feature}
\centering
\begin{tabular}{ccccccc}
\hline
    & All features & Red     & Green        & Blue          & Aspect & DTM \\
IoU & .37          & .31     & \textbf{.27} & .31           & .36    & .33 \\ \hline
    & FlowAccum    & FlowDir & Pcurv        & Slope         & Tcurv  & TWI \\
IoU & .36          & .34     & .34          & \textbf{.26} & .34    & .31 \\ \hline
\end{tabular}
\end{table*}

\subsection{Importance of different features}

To efficiently analyze the importance of different features, we perform inference on "modified" input and report the corresponding IoU score. Specifically for each feature channel, we remove it by replacing with zero values, and check the inference score of remaining features. The lower the score compared to all features remained, the more important the removed feature. Compared to separately training a model for each feature, this "test-time-evaluation" is much less costly. The results are shown in Table \ref{tab:feature}, where green channel and the feature "slope" are clearly the two most important features, and the features "flow accumulation" and "flow direction" are the two least important features. This is also consistent with the visual perception, as we can see from Figure \ref{fig:data} that the orthophoto and the slope are the two most obvious features that align with the ground truth mask.

\subsection{Ambiguity in archaeological labels}

Unlike for natural images commonly used in the computer vision community, the labels for archaeological images are difficult or sometimes impossible to be accurate. There are two main challenges. First, though labels are usually collected from ground surveys by archaeology experts, the ground survey itself can not ensure 100\% accuracy because of the extremely long time gap. Second, the exact boundaries of the ancient landscapes are almost impossible to be accurate, especially for those buried terraces that don't exist on the ground any more. For the first challenge, introducing uncertainty quantification in the labeling phase can greatly improve the quality of the training datasets. For the second challenge, new evaluation metrics may be explored. For example, one can give different weights to pixels with different label confidence. Another example could be introducing object-level metrics like comparing the distance between vectorized central lines of the prediction and the target mask.

\section{Conclusion}

In this work, we present a deep semantic model fusion method for ancient agricultural terrace detection in the Negev desert. We train two types of semantic segmentation models and fuse the predicted probabilities to output the final segmentation map. The experimental results verify the great potential of AI in archeology, but also call for further studies on domain specific characteristics like the ambiguous bundaries of ancient landscapes.

\section*{Acknowledgment}

This work is supported by the Helmholtz Association
through the Framework of Helmholtz AI (grant  number:  ZT-I-PF-5-01) - Local Unit ``Munich Unit @Aeronautics, Space and Transport (MASTr)''.The work of X. Zhu is additionally supported by the German Federal Ministry of Education and Research (BMBF) in the framework of the international future AI lab "AI4EO -- Artificial Intelligence for Earth Observation: Reasoning, Uncertainties, Ethics and Beyond" (grant number: 01DD20001) and by German Federal Ministry for Economic Affairs and Climate Action in the framework of the "national center of excellence ML4Earth" (grant number: 50EE2201C). We thank both IDSI and Helmholtz Information and Data Science Academy (HIDA) for organizing the challenge. 

\printbibliography

\end{document}